%% file: paper.tex
\documentclass{svproc}
\usepackage{url}

\usepackage{multirow}
\usepackage{booktabs}
\usepackage{url}
\usepackage{color,soul}

\begin{document}
\mainmatter              % start of a contribution
\title{%Towards Processing Scholarly Communication Content using Word Embeddings
%Towards Word Embedding Representations for Scholarly Communications
%Word Embeddings in Scholarly Communications: a Trade Off between Corpus Size and Specificity
%A Study on the Use of Distributed Word Representations in Scholarly Communications
Not just about size - A Study on the Role of Distributed Word Representations in the Analysis of Scientific Publications
}
\titlerunning{
Distributed Word Representations in Scientific Publications
}  % abbreviated title (for running head)
\author{Andres Garcia-Silva\inst{1} \and Jose Manuel Gomez-Perez\inst{1}}
\authorrunning{Andres Garcia-Silva and Jose Manuel Gomez-Perez} % abbreviated author list (for running head)
\tocauthor{Andres Garcia-Silva, Jose Manuel Gomez-Perez}
\institute{Expert System, 
C/ Profesor Waksman 10, 28036 Madrid, Spain\\
\email{(agarcia, jmgomez,)@expertsystem.com},\\ WWW home page:
\texttt{http://www.expertsystem.com}
}

\maketitle              % typeset the title of the contribution

\begin{abstract}
The emergence of knowledge graphs in the scholarly communication domain and recent advances in artificial intelligence and natural language processing bring us closer to a scenario where intelligent systems can assist scientists over a range of knowledge-intensive tasks. In this paper we present experimental results about the generation of word embeddings from scholarly publications for the intelligent processing of scientific texts extracted from SciGraph. We compare the performance of domain-specific embeddings with existing pre-trained vectors generated from very large and general purpose corpora. Our results suggest that there is a trade-off between corpus specificity and volume. Embeddings from domain-specific scientific corpora effectively capture the semantics of the domain. On the other hand, obtaining comparable results through general corpora can also be achieved, but only in the presence of very large corpora of well formed text. Furthermore, We also show that the degree of overlapping between knowledge areas is directly related to the performance of embeddings in domain evaluation tasks.

\end{abstract}

\input{introduction}

\input{approach}

\input{experiments}

\input{conclusions}

\section*{Acknowledgments}

We gratefully acknowledge the EU Horizon 2020 program for research and innovation under grant EVER-EST-674907. We also thank Constantino Roman for his contributions to the experimental evaluation of this work.

\bibliographystyle{splncs03}
\bibliography{paper} 

\end{document}

%% file: introduction.tex
\section{Introduction}
In 2017 Springer Nature released the first version of SciGraph\footnote{SciGraph homepage: https://www.springernature.com/gp/researchers/scigraph }, an open linked data graph about publications from the editorial group and cooperating partners. This graph connects funders, research projects, grants, conferences, affiliations, and publications, and in the future it is planned to add citations, patents, and clinical trials. This initiative is a step forward to bring semantics to scholarly publications that contributes to the vision where software agents assist scientists in their research endeavors \cite{gomez2017holistic}. Nevertheless, publication content is still largely text-based, and hence the limitations for the automatic understanding of content, reproducibility of experiments \cite{BECHHOFER2013599}, and knowledge reuse \cite{GomezPerez2017:HumanMachinePartner} remain. 

We envision a scenario where knowledge graphs about Scholarly communications contain semantic metadata about the content of the publication beyond the traditional descriptors used, including keywords and taxonomy categories where articles are placed by authors or editors. This content-based metadata could describe the work hypothesis, conclusions and the approach followed, among others. Therefore, natural language processing (NLP) is a key enabler to extract structured data from scholarly publications that can semantically enrich and shed light on the publication content.

Recently, distributed word representations in the form of dense vectors, known as word embeddings, have been used with great success in NLP tasks such as part-of-speech tagging, chunking, named entity recognition, semantic role labeling and synonym detection \cite{Collobert2008CNNNLP}. Vectors can be learned from large corpora using shallow neural networks \cite{mikolov2013a} or following count-based approaches that perform matrix factorization \cite{pennington2014glove,Levy2014EmbMatrixFac,shazeer2016swivel}. Mikolov et al. \cite{mikolov2013a} showed that embeddings capture semantic relations between words, for example between man and woman, or cities and countries, and syntactic relations based on tenses, singular and plurals, comparatives, superlatives, to name a few, that can be mapped to basic vector operations. 

In this paper we explore the use of word embeddings in the scholarly communications domain through an empirical study. Our goal was to understand whether learning embeddings from a corpus of scientific publications yields better results than using public, pre-trained embeddings generated from very large and general corpora. We learned word embeddings from the publications described in SciGraph. Then, since intrinsic evaluation strategies like word analogies were of limited utility in this case, we used the available metadata contained in the knowledge graph to perform a task-based evaluation consisting of classifying publications along SciGraph's categories. Classifiers were learned through neural networks, including Convolutionals \cite{Collobert2008CNNNLP} (CNN), which have shown good performance in text classification tasks \cite{kim2014convolutional}. In the paper, we also reflect on the ability of CNNs to learn features for the task at hand, which has been proved in computer vision \cite{10.1007/978-3-319-10590-1_53} but still is a matter of debate in text understanding. 

Evaluation results show a trade off between the knowledge specificity of the corpus used to train the word embeddings and its size. In our evaluation task, embeddings from a scientific publication corpus consistently generate classifiers with a top performance that is only matched by classifiers learned from embeddings from very large document corpora such as Common Crawl (http://commoncrawl.org), with 42 billion tokens, or a mix of Wikipedia, news and the UMBC web corpus \cite{UMBCEBIQUITYCORESemanticTextualSimilaritySystems}, with 16 billions tokens. Nevertheless, corpus size seems to lose relevance as the amount of short and informal language it contains increases, e.g. as in Twitter. In addition, we found that embeddings from very specific knowledge fields that are conceptually close tend to perform better in our evaluation task than embeddings from knowledge fields with less overlap. 

The rest of the paper is structured as follows. In section 2 we briefly describe SciGraph content and ontologies. Next, section 3 presents an overview of approaches to generate word embeddings, focusing on FastText and GloVe, that we use in our experiments. In section 4 we introduce the text classification problem and convolutional neural networks to cope with limitations of traditional linear algorithms. In Section 5 we describe our experiments and discuss the results, and finally, in section 6 we present our conclusions.

%% file: approach.tex
\section{A Knowledge Graph for Scholarly Publications}

SciGraph is a linked open data platform for the scientific domain. It comprises information from the complete research process: research projects, conferences, authors and publications, among others. It contains metadata for millions of entities stored in triples. Currently the knowledge graph contains 1 billion facts (triples) about objects of interest to the scholarly domain, distributed over some 85 million entities described using 50 classes and more than 250 properties \cite{HammondPT17:SciGraphModeling}. Currently, most of the knowledge graph is available under CC BY 4.0 License (i.e., attribution) with the exception of abstracts and grant metadata, which are available under CC BY-NC 4.0 License (i.e., attribution and non-comercial)

A core ontology expressed in OWL encodes the semantics of the data in the knowledge graph consisting of 47 classes and 253 properties. Nevertheless, the semantic metadata regarding the publication content is scarce. According to the ontology, just two predicates provide some information at a very high level of abstraction about the article content: i) \textit{sg:hasFieldOfResearchCode} property relates an Australian and New Zealand Standard Research Classification (ANZSRC) Field of Research (FOR) code to a publication, and ii) \textit{sg:hasSubject} property relates a publication to a subject term which describes one of the main topics the publication is about. In addition, text content of publication is limited to titles and abstracts of research articles and book chapters. 

\section{Word Embeddings}
Distributed word representations are based on the distributional hypothesis where words that co-occurr in similar context are considered to have similar (or related) meaning. Word embedding algorithms yield dense vectors so that words with similar distributional context appear in the same region in the embedding space \cite{shazeer2016swivel}. Two main families of algorithms to generate embeddings have been identified \cite{pennington2014glove,levy2015improvingDistSim}: global matrix factorization (count-based) \cite{Levy2014EmbMatrixFac,pennington2014glove,shazeer2016swivel}, and local context window methods (prediction) \cite{mikolov2013a}. Nevertheless, Levy and Goldberg \cite{Levy2014EmbMatrixFac} blurred that distinction, showing that local context window methods like the one proposed by Mikolov et al. \cite{mikolov2013a} are implicitly factorizing a word-context matrix, whose cells are the pointwise mutual information (PMI) of the corresponding word and context pairs, shifted by a global constant. 

%These methods compress the distributional structure of the raw language co-occurrence statistics, yielding compact representations that retain the properties of the original space. The intrinsic quality of the embeddings can be evaluated in two ways. First, words with similar distributional contexts should be near to one another in the embedding space. Second, manipulating the distributional context directly by adding or removing words ought to lead to similar translations in the embedded space, allowing “analogical” traversal of the vector space.

%Models tend to fall into one of two categories: matrix factorization (count-based) or sampling from a sliding window (predicting) \cite{shazeer2016swivel}

%The two main model families for learning word vectors are: 1) global matrix factorization methods, such as latent semantic analysis (LSA) (Deerwester et al., 1990) and 2) local context window methods, such as the skip-gram model of Mikolov et al. (2013c). \cite{pennington2014glove}

\subsection{Word2Vec: Skip-Gram and CBOW}
Mikolov et al. work \cite{mikolov2013a} brought back the research interests to word embeddings in NLP since their approach reduced the computational complexity required to generate word vectors, based on negative sampling and other optimizations. This allowed training with much larger corpora than previous architectures and their evaluation results showed that the vectors encoded semantic and syntactic relations between words that could be calculated with vector operations. 

They proposed two model architectures to compute continuous word vectors from large datasets: Skip-gram and Continuous Bag-of-Words Model (CBOW). In CBOW the words surrounding a central word, i.e., the context, are used to train a log-linear classifier that aims at predicting the word. The Continuous Skip-gram Model is similar to CBOW, but instead of trying to predict a word from its context, it is trained to predict the context of a word. The models were evaluated using semantic and syntactic similarity test sets. The results show that the Skip-gram model significantly outperforms other architectures, specially in the semantic evaluation. 

Levy et al. \cite{levy2015improvingDistSim} showed that much of the performance gains of word embeddings generated with these approaches are due to hyperparameter optimizations and design choices, instead of the embedding algorithms themselves. They also argued that these modifications can be implemented in matrix factorization approaches generating similar performance gains. 

\subsection{FaxtText}
Bojanowski et al.  \cite{bojanowski2016enriching} proposed an evolution of the Skip-Gram model that takes into account n-grams at the character level. The motivation behind it is improving the representation of rare words and taking advantage of the words structure, specially important in morphologically rich languages. This model combines words and subwords in the form of character n-grams. They evaluate their model with nine languages and different word similarity and analogies datasets. The model outperforms Skip-gram and CBOW on almost every dataset. The results also show that computing vectors for out-of-vocabulary words, by summing their n-gram vectors, always obtains equal or better scores than not doing it.

\subsection{GloVe: Global Vectors for Word Representation}
Pennington et al. \cite{pennington2014glove} proposed a model to generate word vectors by training on the nonzero elements of a word coocurrence matrix, instead of training on context windows for each word in the corpus. Their model performs more efficiently by grouping together coocurrence probabilities instead of training in an online manner over the corpus. The model uses a weighted least squares objective. The computational complexity is substantially reduced by training only over the nonzero elements of the coocurrence matrix. Their experiments show that GloVe outperforms Skip-gram and CBOW, among other approaches, while substantially reducing the training time.

\section{Text Classification}
Classifying text documents in one or more classes is a main problem in NLP that has relevant applications, including sentiment analysis, spam detection, email sorting, and vertical search engines that restrict searches to a particular topic \cite{Manning:2008:IIR}. Naive Bayes (NB), their multinomial version (MNB), and Support Vector Machines (SVM) models are frequently used for text classification. Wang et al. \cite{Wang2012LinearClassifiers} showed that NB models perform better in sentiment analysis tasks with short documents, while SVM obtains better results for longer texts. They proposed an approach based on word bigrams to solve the loss of word order present in bag-of-words methods, which is important for sentiment analysis. These algorithms work in highly dimensional space and therefore a feature selection task is required to improve their efficiency and accuracy. However the choice of features is an empirical process, often following trial and error approach, and the features are dependent on the classification task.

\subsection{Convolutional Neural Networks}
Neural network architectures with convolutional layers have been proposed to automatically learn features from text that are relevant to a task at hand, and hence getting rid of manual selection required in the traditional approaches. Convolutional neural networks (CNN) were originally proposed for image recognition and their design is inspired on the visual cortex of the brain. A first layer of simple cells activates with simple elements, like edges or corners, at specific locations, and subsequent layers contain more and more complex cells, which combine simple cells outputs to detect certain shapes, regardless of their position. CNNs are based on convolutional layers \cite{lecun1998gradient} that slide filters (or kernels) across the input data and return the dot products of the elements of the filter and each fragment of the input. The network trains filters to activate with specific features. Stacking several convolutional layers allows feature composition, increasing the level of abstraction as we go from the initial layers to the output.

Single layer CNNs, consisting of a single convolutional layer, pooling, and fully connected neural layers, have been proposed for natural language processing applications. Collobert and Weston \cite{Collobert2008CNNNLP} used a single layer CNN in various NLP-related tasks such as part of speech tagging, named entity recognition, and chuncking, and reached state of the art performance without the need of hand-crafted features. Similarly Kim \cite{kim2014convolutional} used this architecture for text classification and his results improved over the state of the art according to existing benchmarks for sentiment analysis at different granularity levels, detecting subjective and objective sentences, and question classification. 

Multi layer CNN, with more than one convolutional layer, pooling and fully connected neural layers, have been proposed to include information at the character level as a complement to word level information. Dos Santos et al. \cite{dos2014CNNTextClassification} proposed a multi layer CNN with two convolutional layers to analyse sentiments in short texts. Similarly, Zhang et al. \cite{Zhang2015CCN} presented a multi layer convolutional neural network with up to six convolutional layers. Their experiments showed that character-level CNN is an effective method, however its performance depends on many factors, such as dataset size and texts quality among other variables. Convolutional neural networks applied to text classification use word embeddings as input. In some approaches, the CNN architecture learns the embeddings as part of the neural network training \cite{Collobert2008CNNNLP}, while others use pre-trained embeddings \cite{dos2014CNNTextClassification}.

%% file: experiments.tex
\section{Experiments}

So as to produce word embeddings from SciGraph publications we need to gather publication text from the knowledge graph. To this purpose, we query SciGraph for nodes of type \textit{sg:Article} and \textit{sg:BookChapter}, identify research articles and book chapters, and filter them according to publication date in the range 2001 to 2017. For each node, we query its title (\textit{rdfs:label}) and abstract (\textit{sg:abstract}) and keep only publications written in English. In total our dataset consists roughly of 3.2M publications, 1M distinct words, and 886M tokens. In terms of size, it is similar the United Nations corpus \cite{ziemski2016united} (around 600M tokens), which on the other hand is very general.

We use FastText with the Skip-gram algorithm and GloVe to generate embeddings from our dataset. In both cases we generate embeddings with 300 dimensions, truncate the vocabulary at a minimum count of 5 (word frequency), and set the context window size equal to 10. We have faced some issues with the implementation of GloVe when we use the default number of iterations, since some embeddings where produced with null values. In order to address these issues, we decreased the iteration number from 15 to 12. Bear in mind that a lower number of iterations may influence the evaluation of the resulting embeddings.

On the other hand, to compare SciGraph embeddings we use pre-trained embeddings generated with GloVe and FastText learned from very large and general corpora. FastText embeddings where generated from Common Crawl and Wikipedia \cite{grave2018FastTextVectors}, and from Wikipedia exclusively \cite{bojanowski2016enriching}, while GloVe embeddings were learned from Wikipedia, Gigaword, Common Crawl, and Twitter \cite{pennington2014glove}. 
 
\subsection{Analogical Reasoning and Word Similarity}
We initially evaluate our word embeddings through intrinsic evaluation methods \cite{schnabel2015evalMethods}, such as the analogy task \cite{mikolov2013b} and word similarity. The goal of the analogy task is to find x such that the relation x:y resembles a sample relation a:b by operating on the corresponding word vectors. The analogy dataset\footnote{ https://aclweb.org/aclwiki/Google\_analogy\_test\_set\_(State\_of\_the\_art)} contains 19,544 question pairs (8,869 semantic and 10,675 syntactic questions) and 14 types of relations (9 syntactic and 5 semantic). The word similarity evaluation is based on the WordSim353 dataset\footnote{https://aclweb.org/aclwiki/WordSimilarity-353\_Test\_Collection\_(State\_of\_the\_art)}, which contains 353 word pairs with similarity scores assigned by humans that we compare with similarity based on the word embeddings. 

\begin{table}[]
\centering
\begin{tabular}{@{}lrlrrrr@{}}
\toprule
\multirow{2}{*}{Algorithm} & \multicolumn{1}{l}{\multirow{2}{*}{Dimensions}} & \multirow{2}{*}{Corpus} & \multicolumn{3}{c}{Analogy} & \multicolumn{1}{c}{Word Sim.} \\ \cmidrule(l){4-7} 
 & \multicolumn{1}{l}{} & \multicolumn{1}{c}{} & \multicolumn{1}{l}{Sem.} & \multicolumn{1}{l}{Synt.} & \multicolumn{1}{l}{Total} & \multicolumn{1}{l}{Spearman's rho} \\ \midrule
GloVe & 300 & Wiki+Giga (6B) & 77.4 & 67.0 & 71.7 & 0.615 \\
 & 300 & Common Crawl (42B) & \textbf{81.9} & 69.3 & 75.0 & 0.628 \\
FastText & 300 & Wikipedia (3B*) & 77.8 & \textbf{74.9} & \textbf{76.2} & \textbf{0.730} \\ \midrule
GloVe & 300 & SciGraph (886M) & 8.1 & 1.7 & 4.6 & 0.445 \\
FastText & 300 & SciGraph (886M) & 17.1 & 48.5 & 34.3 & 0.587 \\ \bottomrule
\end{tabular}
\caption{Results from the word analogy and similarity tasks}
\label{table:analogy}
\end{table}

Table \ref{table:analogy} reports the accuracy values for SciGraph and pre-trained embeddings in the analogy task and Spearmans's rank correlation coefficient in word similarity. As expected, given the rather small size of the SciGraph corpus (886M tokens) compared to the other sources (number of tokens between 3B and 42B) and the fact that SciGraph focuses on the scientific domain, the performance obtained was significantly lower. A quick look at those benchmarks clearly shows that the SciGraph corpus does not cover all the vocabulary and semantic and syntactic relations that are evaluated in such tasks. However, as we will see below, the fact that SciGraph embeddings do not perform well in this task does not mean that they are not suitable for other tasks focused on the domain from which they were learned. Since word  analogies and similarity are not fit, with the existing benchmarks, to evaluate SciGraph embeddings, we propose an extrinsic evaluation method in the form of a classification task.

\subsection{Classification task}
In SciGraph, each publication has one or more \textit{field of research codes} that classify the documents in 22 categories such as Mathematical Sciences, Engineering or Medical and Health Sciences. Based on this classification scheme, we define a multi-label classification problem to predict one or more categories for each publication through neural networks. The design and configuration of a particular neural network architecture is a complex task that falls out of the scope of this paper. Several approaches try to assist \cite{8109119} data scientist in this task, simplifying the process and helping to select the optimal \cite{DBLP:journals/corr/ZophL16} configuration. In our case, we use two types of neural networks: a regular, fully connected network and a convolutional one. The neural network is composed of an input layer, a fully connected 50-unit neural layer with a ReLU activation function, and an output sigmoid layer. The convolutional network was an out-of-the-box implementation available in Keras with 3 convolutional layers with 128 filters and a 5-element window size, each followed by a max pooling layer, a fully-connected 128-unit ReLU layer and a sigmoid output. 

To evaluate the classifiers we select articles published in 2011 and use ten-fold cross-validation. As baseline, we train a classifier that learns from embeddings generated randomly following a normal distribution. As upper bound we learn a classifier model that learns a new set of word embeddings during training of either the neural and the convolutional networks. The performance of the classifiers is shown in table \ref{table:first-level}. 

The results of the regular neural network show that for this architecture the best classifier is produced from FastText SciGraph embeddings and FastText Wiki+Web+News, although the f-score is far from the upper bound, meaning that there is still room to get better embeddings for this learning algorithm. Looking at the results produced by the convolutional network, we see that all the classifiers have increased their performance to a similar level and the f-measure variation is very close to the upper-bound of 0.79. Also note that the baseline classifier learned from random embeddings has risen its performance to 0.72 and now is closer to the upper bound. 

The fact that, regardless of the corpora used to generate the word embeddings, the convolutional network systematically obtains a performance similar to the best results produced by the upper bound is interesting and shows evidence of the benefits derived from automatic feature selection and composition in text classification. The convolutional network trained on Twitter embeddings is the only case that achieved lower results. This is partially due to the informal language, both vocabulary and grammar, as well as the shorter text found in Twitter, which does not have a significant overlap with the rather formal and specific scientific vocabulary presented in our dataset.

\begin{table}[]
\centering
\begin{tabular}{@{}llrrr@{}}
\toprule
 \multirow{2}{*}{Algorithm} & \multirow{2}{*}{Dataset} & \multicolumn{1}{r}{\multirow{2}{*}{Dim.}} & \multicolumn{2}{c}{F-Score} \\ \cmidrule(l){4-5} 
   &  & \multicolumn{1}{l}{} & NN & CNN \\ \midrule
%Random Uniform &  & 300 & 0.29 & 0.72 \\
Random Normal Dist.&  & 300 & 0.13 & 0.72\\
Classifier &  & 300 & \textbf{0.78} & \textbf{0.79} \\ \midrule
GloVe & Wiki+Giga (6B) & 300 & 0.67 & 0.77\\
 & Common Crawl (42B) & 300 & 0.67 & 0.77 \\
 & Twitter (27B) & 200 & 0.61 & 0.75 \\
FastText & Wiki+Web+News (16B) & 300 & \textbf{0.69} & \textbf{0.78}\\
 & Wiki (3B) & 300 & 0.68 & 0.77 \\ \midrule
GloVe & Scigraph (886M) & 300 & 0.67 & 0.76 \\
FastText & Scigraph (886M) & 300 & \textbf{0.69} & \textbf{0.78}\\ \bottomrule
 %& Scigraph No Ngram & 300 & 0.77 & \textbf{0.69} \\ \bottomrule
\end{tabular}
\caption{Evaluation results for the multi-label classification task}
\label{table:first-level}
\end{table}

\subsection{Fine-Grained Classification}

We further the evaluation of SciGraph embeddings through a fine-grained classification task where we aim at predicting second-level categories in three fields of knowledge: Computer Science, Mathematics, and Chemistry. In addition to the previous embeddings, we generate embeddings for each specific field. We widen the dataset and include publications between 2011 and 2012 so that each second-level category counts on more samples. Finally, we evaluate the multi-label classifiers using ten-fold cross validation and focus on CNNs due to the superior performance showed in the previous experiment. 

Evaluation results (table \ref{table:second-level}) show that embeddings generated from the document corpora of each knowledge field in SciGraph consistently lead to the best classifiers and their performance is very similar to the upper bound. FastText (Wiki+Web+News) embeddings and GloVe (Wiki+Giga) achieve similar f-scores. In this fine-grained classification, we observe that embeddings from the SciGraph general corpus produce average results and do not completely discriminate second-level categories across knowledge fields. This is partially due to differences in corpora size between the general and field-specific cases but also in the semantics captured for each specific word in their corresponding vectors. In the case of the general corpus, the latter is influenced by all the different contexts where a particular word can appear across fields of knowledge. As a consequence, the location in the vector space of the point associated to a particular word is shifted compared to embeddings generated locally to each field.  

\begin{table}[]
\centering
\begin{tabular}{@{}llrrrr@{}}
\toprule
\multirow{2}{*}{Algorithm} & \multirow{2}{*}{Dataset} & \multicolumn{1}{l}{\multirow{2}{*}{Dim.}} & \multicolumn{3}{c}{F-Score} \\ \cmidrule(l){4-6} 
 &  & \multicolumn{1}{l}{} & \multicolumn{1}{l}{Comp. Sci.} & \multicolumn{1}{l}{Math.} & \multicolumn{1}{l}{Chem.} \\ \midrule
%Random Uniform &  & 300 & 0.71 & 0.84 & 0.76 \\
Random Normal &  & 300 & 0.70 & 0.85 & 0.78 \\
Classifier &  & 300 & \textbf{0.79} & \textbf{0.87} & \textbf{0.82} \\ \midrule
GloVe & Wiki+Giga (6B) & 300 & \textbf{0.79} & 0.85 & 0.80 \\
 & Common Crawl (42B) & 300 & \textbf{0.79} & 0.84 & 0.80 \\
 & Twitter (27B) & 200 & 0.73 & 0.84 & 0.78 \\
FastText & Wiki+Web+News (16B) & 300 & \textbf{0.79} & \textbf{0.86} & 0.80 \\
 & Wiki & 300 & 0.77 & 0.85 & 0.77 \\ \midrule
GloVe & Scigraph (886M) & 300 & 0.76 & 0.81 & 0.77 \\
FastText & Scigraph (886M) & 300 & 0.78 & 0.85 & 0.79 \\
%& Scigraph No Ngram & 300 & 0.76 & \textbf{0.86} & 0.78 \\
 & Scigraph (Knowledge-field) & 300 & \textbf{0.79} & \textbf{0.86} & \textbf{0.81} \\ \bottomrule
\end{tabular}
\caption{Evaluation results for fine-grained categories in three knowledge fields}
\label{table:second-level}
\end{table}

In addition, in table \ref{table:specific-fields} we compare the performance of the learned classifiers using only embeddings generated from each knowledge field. In general, embeddings generated from each field train the best classifier for that field. In addition, we note that the level of proximity between knowledge fields appears to be related to these results. For example, embeddings obtained from the Mathematics field perform well in Computer Science and Chemistry, which are both related to Mathematics, and the other way around also holds: embeddings from Computer Science and Chemistry perform well in Mathematics. However, Chemistry embeddings obtain worse results in the Computer Science field and Computer Science embeddings perform worse in the Chemistry category. This could be related to the fact that these fields have less similarities with each other than with Mathematics. Curiously, Computer Science and Chemistry embeddings seem to work even better for Mathematics than for their originating fields, which will need to be further researched. 

\begin{table}[]
\centering
\begin{tabular}{@{}lrrr@{}}
\toprule
\multirow{2}{*}{Embeddings} & \multicolumn{3}{c}{F-Score} \\ \cmidrule(l){2-4} 
 & \multicolumn{1}{l}{Computer Science} & \multicolumn{1}{l}{Mathematics} & \multicolumn{1}{l}{Chemistry} \\ \midrule
Computer Science & \textbf{0.79} & \textbf{0.86} & 0.77 \\
Mathematics & 0.77 & \textbf{0.86} & 0.80 \\
Chemistry & 0.75 & 0.85 & \textbf{0.81} \\ \bottomrule
\end{tabular}
\caption{Comparison between embeddings generated from specific fields}
\label{table:specific-fields}
\end{table}

%% file: conclusions.tex
\section{Conclusions}
This paper presents experimental results related to the generation and evaluation of word embeddings from scholarly communications in the scientific domain, leveraging SciGraph content and metadata. We compare the resulting domain-specific embeddings with pre-trained vectors generated from large and general purpose corpora in intrinsic and extrinsic tasks. 

We show that intrinsic evaluation methods like word analogy and word similarity are not a reliable benchmark for embeddings learned from scientific corpora, mainly due to the mismatch in vocabulary and semantics between the scientific corpus and the evaluation dataset. This kind of findings should increase awareness on the need for larger (or domain-specific) word embeddings intrinsic evaluation benchmarks. We then followed on to conduct an extrinsic evaluation in the form of a domain-dependent classification task, at different granularity levels, for scientific publications. The evaluation shows that our classifiers make the most of embeddings generated through FastText both from a corpus of scientific publications (886 million of tokens) and from a much larger mixture of Wikipedia, Web content, and News (16 billions of tokens). 

Our results pose a trade-off between corpus size and specificity in domain-dependent tasks. That is, for scholarly communications, embeddings learned from focused corpora produce similar results that embeddings generated from much larger and general corpora with many billions of tokens. Furthrmore, we noticed that all the classifiers learned through convolutional neural networks were closer to the upper bound, indicating that the features learned by the convolutional network are more expressive than pre-trained word embeddings. In addition we showed that embeddings generated from specific knowledge fields perform well in classification tasks of semantically related knowledge fields such as Computer Science and Mathematics, and not so well where the knowledge fields are not so related such as Computer Science and Chemistry. 

We expect that this work lays a foundation for the future use of embeddings in NLP tasks applied to scientific publications. For example, to enhance and curate existing knowledge graphs, such as SciGraph itself, with metadata about the publication content so that not only accessory, but also content-wise structured metadata is available for software applications. Related work in this direction involves merging embeddings and Expert System's Cogito knowledge graph \footnote{http://www.expertsystem.com/products/cogito-cognitive-technology/} in a single vector space \cite{DenauxG17:vecsigrafo}, which showed improvements in word similarity evaluations with respect to traditional word embeddings and other NLP downstream tasks.

%% file: paper.bbl
\begin{thebibliography}{10}
\providecommand{\url}[1]{\texttt{#1}}
\providecommand{\urlprefix}{URL }

\bibitem{BECHHOFER2013599}
Bechhofer, S., Buchan, I., Roure, D.D., Missier, P., Ainsworth, J., Bhagat, J.,
  Couch, P., Cruickshank, D., Delderfield, M., Dunlop, I., Gamble, M.,
  Michaelides, D., Owen, S., Newman, D., Sufi, S., Goble, C.: Why linked data
  is not enough for scientists. Future Generation Computer Systems  29(2),  599
  -- 611 (2013), special section: Recent advances in e-Science

\bibitem{bojanowski2016enriching}
Bojanowski, P., Grave, E., Joulin, A., Mikolov, T.: Enriching word vectors with
  subword information. arXiv preprint arXiv:1607.04606  (2016)

\bibitem{Collobert2008CNNNLP}
Collobert, R., Weston, J.: A unified architecture for natural language
  processing: Deep neural networks with multitask learning. In: Proceedings of
  the 25th International Conference on Machine Learning. pp. 160--167. ICML
  '08, ACM, New York, NY, USA (2008),
  \url{http://doi.acm.org/10.1145/1390156.1390177}

\bibitem{DenauxG17:vecsigrafo}
Denaux, R., G{\'{o}}mez{-}P{\'{e}}rez, J.M.: Towards a vecsigrafo: Portable
  semantics in knowledge-based text analytics. In: Joint Proceedings of the
  International Workshops on Hybrid Statistical Semantic Understanding and
  Emerging Semantics, and Semantic Statistics co-located with 16th
  International Semantic Web Conference, HybridSemStats@ISWC 2017, Vienna,
  Austria October 22nd, 2017 (2017),
  \url{http://ceur-ws.org/Vol-1923/article-04.pdf}

\bibitem{8109119}
Desell, T.: Developing a volunteer computing project to evolve convolutional
  neural networks and their hyperparameters. In: 2017 IEEE 13th International
  Conference on e-Science (e-Science). pp. 19--28 (Oct 2017)

\bibitem{GomezPerez2017:HumanMachinePartner}
Gomez-Perez, J.M., Palma, R., Garcia-Silva, A.: Towards a human-machine
  scientific partnership based on semantically rich research objects. In: 2017
  IEEE 13th International Conference on e-Science (e-Science). pp. 266--275
  (Oct 2017)

\bibitem{gomez2017holistic}
Gomez-Perez, J.M., Denaux, R., Garcia-Silva, A., Palma, R.: A holistic approach
  to scientific reasoning based on hybrid knowledge representations and
  research objects. In: Proceedings of Workshops and Tutorials of the 9th
  International Conference on Knowledge Capture (K-CAP2017). pp. 47--49 (2017)

\bibitem{grave2018FastTextVectors}
Grave, E., Bojanowski, P., Gupta, P., Joulin, A., Mikolov, T.: Learning word
  vectors for 157 languages. In: Proceedings of the International Conference on
  Language Resources and Evaluation (LREC 2018) (2018)

\bibitem{HammondPT17:SciGraphModeling}
Hammond, T., Pasin, M., Theodoridis, E.: Data integration and disintegration:
  Managing springer nature scigraph with shacl and owl. In: Nikitina, N., Song,
  D., Fokoue, A., Haase, P. (eds.) International Semantic Web Conference
  (Posters, Demos and Industry Tracks). CEUR Workshop Proceedings, vol. 1963.
  CEUR-WS.org (2017),
  \url{http://dblp.uni-trier.de/db/conf/semweb/iswc2017p.html#HammondPT17}

\bibitem{UMBCEBIQUITYCORESemanticTextualSimilaritySystems}
Han, L., Kashyap, A.L., Finin, T., Mayfield, J., Weese, J.: {UMBC
  EBIQUITY-CORE: Semantic Textual Similarity Systems}. In: Proceedings of the
  Second Joint Conference on Lexical and Computational Semantics. Association
  for Computational Linguistics (June 2013)

\bibitem{kim2014convolutional}
Kim, Y.: Convolutional neural networks for sentence classification. In:
  Proceedings of the 2014 Conference on Empirical Methods in Natural Language
  Processing, {EMNLP} 2014, October 25-29, 2014, Doha, Qatar, {A} meeting of
  SIGDAT, a Special Interest Group of the {ACL}. pp. 1746--1751 (2014),
  \url{http://aclweb.org/anthology/D/D14/D14-1181.pdf}

\bibitem{lecun1998gradient}
LeCun, Y., Bottou, L., Bengio, Y., Haffner, P.: Gradient-based learning applied
  to document recognition. Proceedings of the IEEE  86(11),  2278--2324 (1998)

\bibitem{Levy2014EmbMatrixFac}
Levy, O., Goldberg, Y.: Neural word embedding as implicit matrix factorization.
  In: Proceedings of the 27th International Conference on Neural Information
  Processing Systems - Volume 2. pp. 2177--2185. NIPS'14, MIT Press, Cambridge,
  MA, USA (2014), \url{http://dl.acm.org/citation.cfm?id=2969033.2969070}

\bibitem{levy2015improvingDistSim}
Levy, O., Goldberg, Y., Dagan, I.: {Improving Distributional Similarity with
  Lessons Learned from Word Embeddings}. Transactions of the ACL  3(0),
  211--225 (2015)

\bibitem{Manning:2008:IIR}
Manning, C.D., Raghavan, P., Sch\"{u}tze, H.: Introduction to Information
  Retrieval. Cambridge University Press, New York, NY, USA (2008)

\bibitem{mikolov2013b}
Mikolov, T., Sutskever, I., Chen, K., Corrado, G., Dean, J.: {Distributed
  Representations of Words and Phrases and their Compositionality.} In: NIPS
  (2013)

\bibitem{mikolov2013a}
Mikolov, T., Chen, K., Corrado, G., Dean, J.: Efficient estimation of word
  representations in vector space. CoRR  abs/1301.3781 (2013),
  \url{http://arxiv.org/abs/1301.3781}

\bibitem{pennington2014glove}
Pennington, J., Socher, R., Manning, C.D.: {Glove: Global vectors for word
  representation.} In: EMNLP. vol.~14, pp. 1532--1543 (2014)

\bibitem{dos2014CNNTextClassification}
dos Santos, C., Gatti, M.: Deep convolutional neural networks for sentiment
  analysis of short texts. In: Proceedings of COLING 2014, the 25th
  International Conference on Computational Linguistics: Technical Papers. pp.
  69--78 (2014)

\bibitem{schnabel2015evalMethods}
Schnabel, T., Labutov, I., Mimno, D., Joachims, T.: {Evaluation methods for
  unsupervised word embeddings}. In: EMNLP. pp. 298--307. ACL (2015)

\bibitem{shazeer2016swivel}
Shazeer, N., Doherty, R., Evans, C., Waterson, C.: {Swivel: Improving
  Embeddings by Noticing What's Missing}. arXiv preprint  (2016)

\bibitem{Wang2012LinearClassifiers}
Wang, S., Manning, C.D.: Baselines and bigrams: Simple, good sentiment and
  topic classification. In: Proceedings of the 50th Annual Meeting of the
  Association for Computational Linguistics: Short Papers - Volume 2. pp.
  90--94. ACL '12, Association for Computational Linguistics, Stroudsburg, PA,
  USA (2012), \url{http://dl.acm.org/citation.cfm?id=2390665.2390688}

\bibitem{10.1007/978-3-319-10590-1_53}
Zeiler, M.D., Fergus, R.: Visualizing and understanding convolutional networks.
  In: Fleet, D., Pajdla, T., Schiele, B., Tuytelaars, T. (eds.) Computer Vision
  -- ECCV 2014. pp. 818--833. Springer International Publishing, Cham (2014)

\bibitem{Zhang2015CCN}
Zhang, X., Zhao, J., LeCun, Y.: Character-level convolutional networks for text
  classification. In: Proceedings of the 28th International Conference on
  Neural Information Processing Systems - Volume 1. pp. 649--657. NIPS'15, MIT
  Press, Cambridge, MA, USA (2015),
  \url{http://dl.acm.org/citation.cfm?id=2969239.2969312}

\bibitem{ziemski2016united}
Ziemski, M., Junczys-Dowmunt, M., Pouliquen, B.: {The united nations parallel
  corpus v1. 0}. In: Language Resource and Evaluation (2016)

\bibitem{DBLP:journals/corr/ZophL16}
Zoph, B., Le, Q.V.: Neural architecture search with reinforcement learning.
  CoRR  abs/1611.01578 (2016), \url{http://arxiv.org/abs/1611.01578}

\end{thebibliography}
